\documentclass{article}
\usepackage{spconf,amsmath,graphicx}

\usepackage{enumitem}
\setlist{nosep, leftmargin=14pt}

\usepackage{mwe} 

\title{Auxiliary CycleGAN-guided Task-Aware Domain Translation \\ from Duplex to Monoplex IHC Images}
\name{
  \begin{tabular}{@{}c@{}}
  Nicolas Brieu$^{\star}$ 
  \qquad
  Nicolas Triltsch$^{\star}$ 
  \qquad
  Philipp Wortmann$^{\star}$ 
  \qquad
  Dominik Winter$^{\star}$ \\
  \qquad
  Shashank Saran$^{\dagger}$ 
  \qquad
  Marlon Rebelatto$^{\dagger}$
  \qquad
  G{\"u}nter Schmidt$^{\star}$ 
  \end{tabular}
}

\address{
$^{\star}$ AstraZeneca Computational Pathology, Oncology Research and Development, Munich, Germany \\ 
$^{\dagger}$ AstraZeneca Translational Pathology, Oncology Research and Development, Gaithersburg, US
}

\begin{document}
\maketitle
\begin{abstract}
Generative models enable the translation from a source image domain where readily trained models are available to an unseen target domain. 
While Cycle Generative Adversarial Networks (CycleGANs) are commonly used for unpaired translation, the associated cycle consistency constrain assumes an invertible mapping between the two domains. 
However, this assumption does not hold true for the translation between images stained with chromogenic monoplex and duplex immunohistochemistry (IHC) assays. 
To address this, we introduce of a novel training design which leverages a set of immunofluorescence (IF) images as an auxiliary and unpaired image domain to guide the direct translation from duplex to monoplex IHC. 
The results, both quantitative and qualitative, on a downstream segmentation task demonstrate the advantages of the proposed method over multiple baseline methods.
\end{abstract}

\section{Introduction}
Numerous generative networks for domain transfer have been explored and recently applied to the field of computational pathology, particularly those based on generative adversarial networks (GANs) \cite{zhu2017} and diffusion models \cite{jose2021generative}. While diffusion models often outperform GANs on image synthesis tasks, they typically suffer from significantly longer inference times, making GANs still prevalent in practical applications. Computational efficiency is crucial in a field where training datasets and whole slide images (WSI) can consists of thousands of fields of views (FOVs) and several billions of pixels, respectively. CycleGANs are computationally efficient models that, unlike methods such as Pix2Pix \cite{isola2017}, can be trained on unpaired sets of images and have seen applications in stain normalization \cite{shaban2019, mahapatra2020}, stain domain transfer \cite{brieu2019, brieu2022stain} and data augmentation \cite{wagner2021structure, berijanian2023, gour2024histopathological}. 

In this work, we consider a labeled dataset of monoplex immunohistochemistry (IHC) images stained with a DAB cell membrane marker (e.g. HER2) and a hematoxylin (HTX) nuclear counter stain (source domain) as well as an existing segmentation model \cite{kapil2024} trained on this source dataset. 
The objective is to extend the use of this segmentation model to a target image domain consisting of duplex IHC images stained with a DAB membrane stain (e.g. HER2), a purple eosin-like nuclear stain (e.g. Ki67) and a HTX counter stain. 
The source and target domains are unpaired. 
By cascading the existing segmentation model after a domain translation model from the target duplex IHC domain to the source monoplex IHC domain, we obtain a segmentation model directly usable on the target duplex IHC domain. 
However, we observe that the assumption of a bijective mapping between the source and target domains, which is required by the cycle consistency loss used in CycleGAN, does not hold in the particular case of translation between duplex and monoplex IHC images. 
Multiple nuclear signals (eosin-like and HTX) in the first domain are mapped to a single nuclear signal  (HTX) in the second domain and vice versa, leading to ambiguity.

To bypass this limitation, we leverage and extend our recently introduced ReStainGAN model \cite{Winter2024}, which builds on CycleGAN to allow for easy manipulation of stain representations in IHC stained images using an auxiliary and unpaired immunofluorescence (IF) image domain. 
By selecting a well-chosen combination of markers in the IF domain, the method enables the disentanglement of the convoluted stain components in the IHC domain into separate channels in the IF domain. 
This allows stain manipulation in the IHC domain to be formulated as simple mathematical operations in the IF domain. 
Back-translation of the manipulated images to the original IHC domain results in \textit{in-silico} images mimicking the target IHC domain of interest. 
In this work, we extend the ReStainGAN model from monoplex to duplex and learn the CycleGAN-based bijective mapping between duplex IHC images in the source domain and a set of unpaired IF images with e.g. DAPI, HER2 and Ki67 markers. 
The different marker expressions are then manipulated in the IF domain to transform a true duplex IHC image into a synthetic monoplex IHC image. Following a pix2pix architecture \cite{isola2017}, the resulting synthetic monoplex IHC image then guides the direct paired translation by an independent generative adversarial network from the duplex IHC domain to the monoplex IHC domain. The primary contribution of this work is the introduction of an auxiliary CycleGAN to circumvent the non-invertible mapping associated with the transformation of interest. 

\section{Methods}
Let's denote domain $A$ a set of duplex IHC images $(x_a)_{x_A\in A}$ with DAB membrane marker (e.g. HER2), nuclear eosin-like marker (e.g. Ki67) and HTX counter stain. Domain $B$ is a set of monoplex IHC images $(x_B)_{x_B\in B}$ with DAB membrane marker and HTX counter stain. Finally, domain $C$ is a set of IF stained images $(x_C)_{x_C\in C}$ consisting of a DAPI channel, a membrane marker channel (e.g. HER2) and a nuclear marker channel (e.g. Ki67). All domains are unpaired. 

Due to the equivalence between the HTX-eosin-DAB (HED) components of domain $A$ and the IF components of domain $C$, the bijective mapping between the two stain domains can be learned by a CycleGAN with two generator models $\mathcal{G}_{AC}$ and $\mathcal{G}_{CA}$ performing duplex IHC to IF and IF to duplex IHC translation, respectively. Two losses based on color deconvolution by Bayesian K-Singular Value Decomposition (BKSVD) \cite{perez2022bayesian} are added to the standard adversarial and cycle consistency losses to avoid cycle-consistent but non structure-preserving errors \cite{brieu2022stain}. 

Given a sample $x_A$ from the source domain $A$, the associated separated HED components yielded by its transformation to the IF domain by the generator $\mathcal{G}_{AC}$ read as $x_{HED}=\mathcal{G}_{AC}(x_A):=(x_{|H}, x_{|E}, x_{|D})$. Building on our previous two-marker ReStainGAN model \cite{Winter2024}, the three HED components are modified by a so-called restaining function $\kappa()$ and transformed back to the original RGB colorspace, yielding the transformed IHC image $x_A'$:
\begin{equation}
\label{eq:restaingan}
x_A'=f_{AB}(x_A)=\mathcal{G}_{CA}\circ\kappa\circ\mathcal{G}_{AC}(x_A), 
\end{equation}
with the restaining function defined as:
\begin{equation}
\label{eq:hed}
\begin{aligned}
\kappa_{\alpha}(x_{HED})_{|H} &= \alpha_{hh} x_{|H} + \alpha_{eh} x_{|E} + \alpha_{dh} x_{|D}\\
\kappa_{\alpha}(x_{HED})_{|E} &= \alpha_{he} x_{|H} + \alpha_{ee} x_{|E} + \alpha_{de} x_{|D}\\
\kappa_{\alpha}(x_{HED})_{|D} &= \alpha_{hd} x_{|H} + \alpha_{ed} x_{|E} + \alpha_{dd} x_{|D}.
\end{aligned}
\end{equation}
Keeping the HTX and DAB signal unchanged and transforming the eosin staining into HTX yields $\alpha_{hh}=1.0$, $\alpha_{dd}=1.0$ and $\alpha_{eh}=0.5$ with all other values set to 0. Note that $\alpha_{eh}$ is experimentally set to a value lower than 1 to account for the observed lower saturation of the Hematoxylin marker compared to the functional eosin-like nucleus staining. 

\begin{figure}[t!]
\centering
\includegraphics[width=1.0\linewidth]{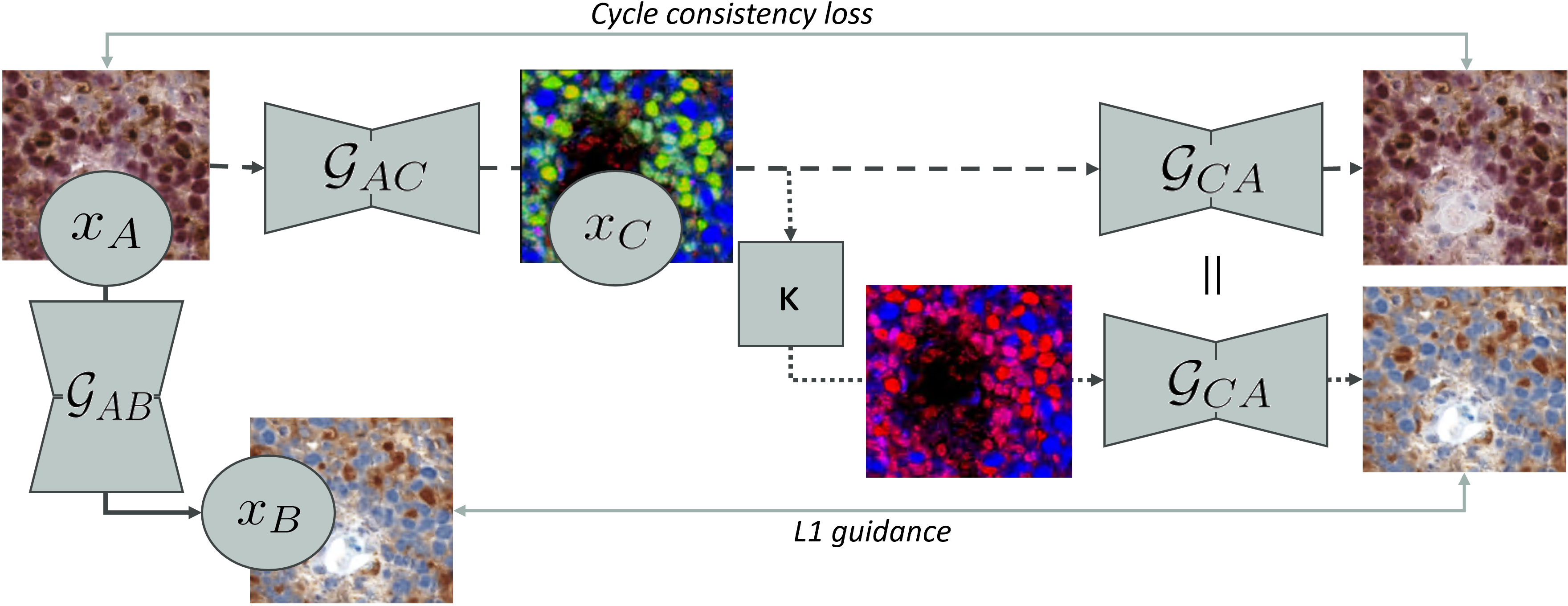}
\caption{Auxiliary CycleGAN $\mathcal{G}_{AC}$ and $\mathcal{G}_{CA}$ between duplex IHC (A) and IF (C) image domain for the task-aware guidance of the GAN-based translation $\mathcal{G}_{AB}$ of duplex IHC images to the monoplex IHC image domain (B).}
{\label{fig:method}}
\end{figure}

Given a segmentation model $S_B$ trained on a subset of labeled images in domain $B$, we aim at finding the generative function $\mathcal{G}_{AB}$ performing the direct translation from domain $A$ to domain $B$ which would enable the definition of the cascaded model $S_B\circ \mathcal{G}_{AB}$ and its use on domain $A$. To this end, we optimize the parameters of the generative model $\mathcal{G}_{AB}$ via the minimization of the following loss function:
\begin{equation}
\label{eq:loss_ga2b}
\mathcal{L}_\mathcal{G} = \mathcal{L}_\mathcal{G}^{AB} + \lambda_0 \mathcal{L}^{AB}_{g},
\end{equation}
where $\mathcal{L}_\mathcal{G}^{AB}$ denotes the least square objective of the GAN \cite{mao2017} and where $\mathcal{L}^{AB}_{g}$ is a guidance loss function defined by:
\begin{equation}
\label{eq:ustain}
\mathcal{L}^{AB}_{g}(x_A) = ||\mathcal{G}_{AB}(x_A)-f_{AB}(x_A)||_1.
\end{equation}
$f_{AB}$ is defined by Equation~\eqref{eq:restaingan} and performs the task-aware translation of the duplex IHC image $x_A$ into an \textit{in-silico} IHC-monoplex image. The CycleGAN-based translation between the duplex IHC and IF domains by $\mathcal{G}_{AC}$ and $\mathcal{G}_{CA}$ is auxiliary to the direct translation from the duplex IHC to the monoplex IHC domains by $\mathcal{G}_{AB}$. The generators between the stain domains are illustrated in Figure~\ref{fig:method}. The respective discriminators are not represented for simplicity. Equation~\eqref{eq:ustain} becomes:
\begin{equation}
\label{eq:ustain2}
\mathcal{L}^{AB}_{g}(x_A) = ||\mathcal{G}_{AB}(x_A)-\mathcal{G}_{CA}\circ \kappa \circ \mathcal{G}_{AC}(x_A)||_1.
\end{equation}
The set of monoplex IHC images being a subset of the duplex IHC images, images from both the duplex and monoplex IHC domains are used for training $\mathcal{G}^{AB}$. Also, since monoplex IHC images are expected to be transformed into themselves by $\mathcal{G}^{AB}$, an identity loss $\mathcal{L}^B_{id}$ is defined as follows $\mathcal{L}^B_{id}(x_B) = ||\mathcal{G}_{AB}(x_B)-x_B)||_1$ and included in the training.

\begin{figure*}[t!]
\includegraphics[width=0.925\linewidth]{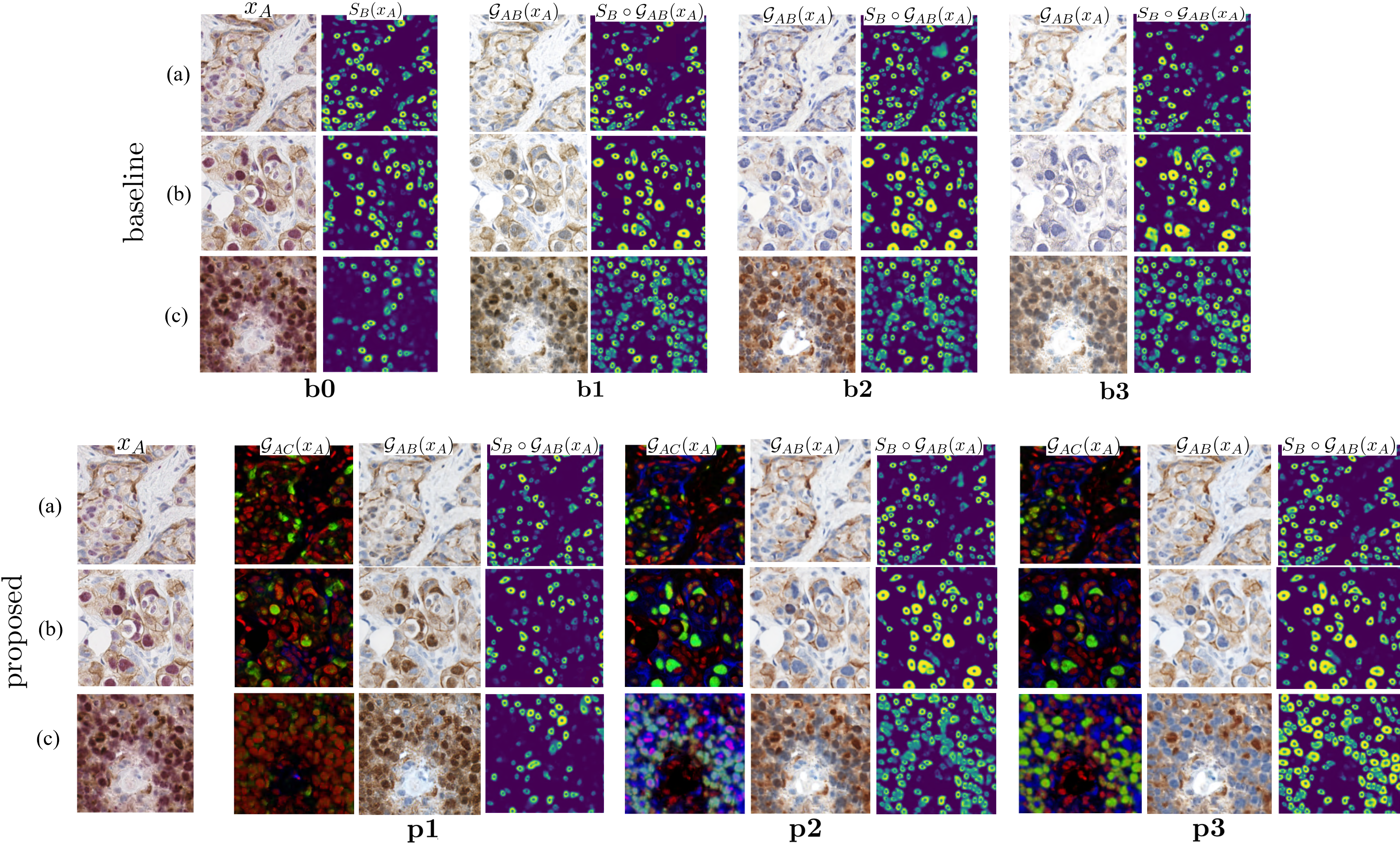}
\caption{Qualitative results on representative duplex IHC images $x_A$, displayed at the foremost left column of each row. The synthetic monoplex IHC images $\mathcal{G}_{AB}(x_A)$ and the corresponding posterior maps $S_B\circ\mathcal{G}_{AB}(x_A)$ associated to the downstream nucleus segmentation task are shown for each baseline method $\textbf{b1-3}$. Predicted centers appear as holes in the posterior maps. For each method $\textbf{p1-3}$ the synthetic images $\mathcal{G}_{AC}(x_A)$ generated by the auxiliary CycleGAN are additionally shown first. }
\vspace{-2mm} 
{\label{fig:res_qual}}
\end{figure*}

The use of an auxiliary image domain $C$ where the stain components are well separated enables the introduction of two supervised losses to complement the associated unpaired and otherwise unsupervised translation between domains $A$ and $C$. The first supervised loss constrains the non-existence of the eosin-like component in the synthetic IF images resulting from the translation by the generative network $\mathcal{G}_{AC}$ of the monoplex IHC images in domain $B$ from domain $A$ to domain $C$: $\mathcal{L}^{BC}_{E}(x_B) = ||\mathcal{G}_{AC}(x_B)_{|E}||_1$. The second leverages a small amount of labeled samples - obtained by coarse, sparse and therefore time effective selection of eosin and DAB pixels in the duplex IHC images. Given $x_A$ an duplex IHC patch as well as $\mathcal{M}_A^E$ and $\mathcal{M}_A^D$ the masks corresponding to the selected pixels, the separation of the eosin-like and DAB components in the synthetic IF images is favored by the two following losses:
\vspace{-2mm} 
\begin{equation} 
\label{eq:sup}
\begin{aligned}
\mathcal{L}^{AC}_{sE} &= \frac{1}{|\mathcal{M}_A^E|}\sum_{p \in \mathcal{M}_A^E} \frac{\left(\mathcal{G}_{AC}(x_A)_{|E}\right)^2}{||\mathcal{G}_{AC}(x_A))||_1},\\ 
\mathcal{L}^{AC}_{sD} &= \frac{1}{|\mathcal{M}_A^D|}\sum_{p \in \mathcal{M}_A^D} \frac{\left(\mathcal{G}_{AC}(x_A)_{|D}\right)^2}{||\mathcal{G}_{AC}(x_A))||_1},
\end{aligned}
\vspace{-2mm} 
\end{equation}
which constrain the generative network $\mathcal{G}_{AC}$ such that the eosin-like (resp. DAB) component in the translated \textit{in-silico} IF image $\mathcal{G}_{AC}(x_A)$ is high in absolute and relative values in the sparsely annotated set of eosin (resp. DAB) pixels.

The inclusion of the standard losses for CycleGAN training, of the  stain isolation losses and of the newly introduced null eosin and supervised losses yields the following extension of the generator loss initially defined in Equation~\eqref{eq:loss_ga2b}: 
\vspace{-1mm} 
\begin{equation}
\label{eq:loss_g}
\begin{aligned}
\mathcal{L}_{\mathcal{G}} &= \mathcal{L}_\mathcal{G}^{AC} + \mathcal{L}_\mathcal{G}^{CA} + \lambda_1 \mathcal{L}^{AC}_{cycle} \\
  & + \lambda_2 \left(\mathcal{L}_{HED}^{AC} + \mathcal{L}_{RGB}^{CA}\right) \\
  &+ \lambda_3 \mathcal{L}^{BC}_E + \lambda_4 \left(\mathcal{L}^{AC}_{sE} + \mathcal{L}^{AC}_{sD}\right) \\
	&+ \mathcal{L}_\mathcal{G}^{AB} + \lambda_0 \mathcal{L}^{AB}_{g},
\end{aligned}
\end{equation}
The first three lines correspond to the training of the auxiliary CycleGAN between unpaired domains $A$ and $C$ including its extension to stain-isolation-based and supervised guidance respectively. The fourth line corresponds to the direct translation from domain $B$ to domain $A$.

\section{Results}

\begin{figure*}[t!]
  \includegraphics[width=0.33\linewidth]{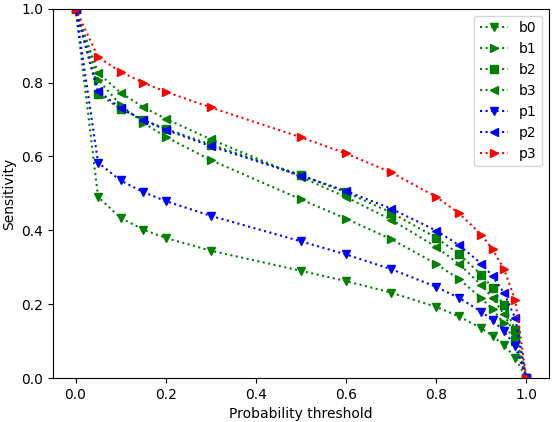}
  \includegraphics[width=0.33\linewidth]{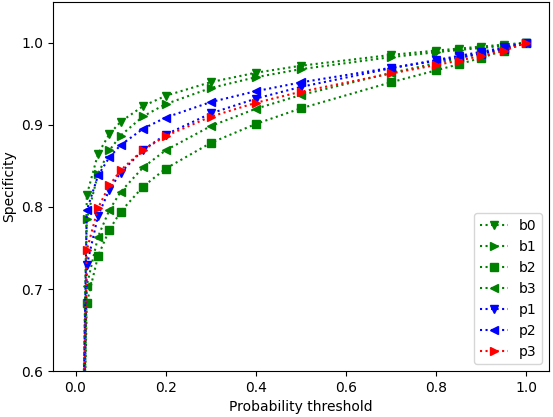}
  \includegraphics[width=0.33\linewidth]{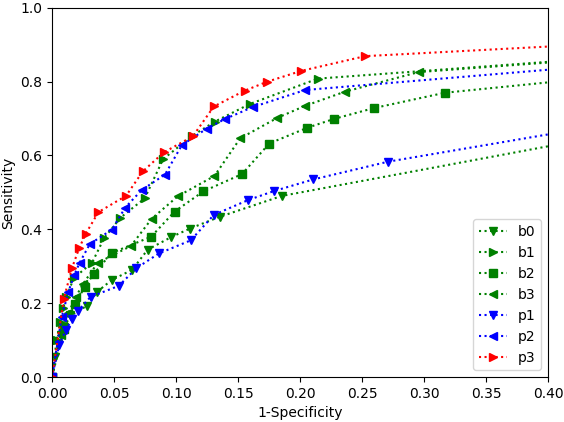}
\vspace{-2mm} 
\caption{Left and middle: Sensitivity and specificity curves for different threshold values applied on the predicted nucleus posterior. Left: corresponding Receiver Operating Characteristic (ROC) curves. }
\vspace{-2mm} 
\label{fig:res_quant}
\end{figure*}

The duplex IHC dataset consists of 16 WSIs from which 6.9k unlabeled patches of $256\times256$px and $0.5 {\mu}$/px resolution are generated. Additionally, a set of 18 FOVs is selected on these images and a sparse amount of eosin-like and DAB-stained pixels manually labeled, yielding a small set of 200 partially labeled patches. The test set consists of 8 FOVs from 4 WSIs and manually annotated for nucleus segmentation of epithelium cells, totaling 1362 nuclei. The unpaired monoplex IHC and IF datasets consists of 35k and 4.5k unlabeled patches, respectively, of the same size and resolution as the duplex IHC dataset. 
All generator and discriminator models are optimized using Adam optimizer with learning rates of 0.0001 and 0.0005, respectively. The CycleGAN for translation between IF and IHC-duplex is pre-trained for 15k iterations to ensure that an already realistic reference image is used in the restain guidance loss $\mathcal{L}^{AB}_{g}$ while the null-eosin $\mathcal{L}^{BC}_E$ and supervision $\mathcal{L}^{AC}_{sE}$, $\mathcal{L}^{AC}_{sD}$ losses are applied only after 2.5k iterations to allow for the initialization of $\mathcal{G}_{AC}(x_A)$. The full network now including $\mathcal{G}_{AB}$ is trained for a total of 30k iterations. The following parameter values are experimentally determined and fixed through the experiments: $\lambda_0 = \lambda_1 = 10$. This setup ensures that effective translations are learned between the different stain domains. Ablation studies are performed on the stain isolation ($\lambda_2$), null-eosin ($\lambda_3$) and supervised losses ($\lambda_4$), yielding the configurations: 
\begin{itemize}
\item \textbf{p1}: $\lambda_2=0, \lambda_3=0, \lambda_4=0$,
\item \textbf{p2}: $\lambda_2=5, \lambda_3=10, \lambda_4=0$,
\item \textbf{p3}: $\lambda_2=5, \lambda_3=10, \lambda_4=0.5$.
\end{itemize}
Qualitative and quantitative results are reported in Figures \ref{fig:res_qual} and \ref{fig:res_quant} as well as in Table \ref{tab:results} for the above parameter configurations \textbf{p1-3} of the proposed method as well as for the following four baseline approaches: 
\begin{itemize}
\item \textbf{b0}: the direct use of the segmentation model trained on IHC-monoplex images on IHC-duplex input images;
\item \textbf{b1}: the use of BKSVD color deconvolution-based restaining instead of $\mathcal{G}_{AB}$ for translating the duplex IHC into a monoplex IHC;
\item \textbf{b2}: the translation of the input IHC-duplex image into IHC-monoplex using the non-bijective CycleGAN directly defined between these two domains;
\item \textbf{b3}: the use of BKSVD color deconvolution-based restaining instead of the proposed auxiliary CycleGAN for guiding the training of $\mathcal{G}_{AB}$.
\end{itemize}

We perform a quantitative evaluation on the downstream task of nucleus segmentation. Sensitivity is computed as the percentage of annotated nucleus pixels for which the estimated nucleus probability exceeds a given threshold, and specificity as the percentage of annotated background pixels for which the estimated nucleus probability stays below a given threshold. This results in a Receiver Operating Characteristic (ROC) curve plot, displayed along with the sensitivity and specificity curve plots in Figure \ref{fig:res_quant}. The corresponding Area Under the Curve (AUC) values are reported in Table \ref{tab:results}. 

The proposed approach $\textbf{p3}$, yields the highest ROC-AUC of $0.8621$, which is to be compared to a ROC-AUC of $0.6702$ ($-22\%$) obtained by directly using the segmentation model trained on monoplex IHC on duplex IHC without translation (\textbf{b0}), and to a ROC-AUC of $0.8324$ ($-3.5\%$) obtained by using BKSVD color deconvolution-based restaining (\textbf{b1}). 
Both ablation study-models, \textbf{p1} and \textbf{p2}, yield lower ROC-AUC compared to proposed \textbf{p3}. 
The proposed method \textbf{p3} yields a significant increase in sensitivity ($+29.8\%$) for only a small decrease in specificity ($-2.8\%$) compared to the best baseline method \textbf{b1} in terms of ROC. The lack of sensitivity in the latter method can be observed in Figure~\ref{fig:res_qual}(c) and is explained by its inability to correctly transform increasingly saturated nuclei. This evaluation demonstrates the effectiveness and superiority of the proposed approach in achieving higher segmentation accuracy and sensitivity.

\begin{table}[t]
  \begin{center}
  {\small{
	\vspace{-1mm}} 
  \begin{tabular}{c|| c| c| c|}
    {Method} & {Sensitivity} & {Specificity} & {ROC} \\
		\hline
		\hline
    \textbf{b0} & 0.2989 & 0.9480 & 0.6702\\
		\textbf{b1} & 0.4833 & 0.9415 & 0.8323\\
		\textbf{b2} & 0.5279 & 0.8928 & 0.7749\\
		\textbf{b3} & 0.5288 & 0.9077 & 0.8107\\
		\hline
		\textbf{p1} & 0.3718 & 0.9187 & 0.6859\\
		\textbf{p2} & 0.5349 & 0.9295 & 0.8207\\
		\textbf{p3} & 0.6272 & 0.9156 & \textbf{0.8621}\\	
\end{tabular}
\vspace{-2mm}
}
\end{center}
\caption{Area Under the curve (AUC) values computed for the sensitivity, specificity and ROC plots shown in Figure~\ref{fig:res_quant} and associated with the different baseline \textbf{b0-1}, proposed \textbf{p3} and corresponding degraded \textbf{p1-2} methods. \vspace{-1mm}} 
\label{tab:results}
\end{table}

\vspace{-1mm}
\section{Conclusion}
We propose a novel inference-time stain translation approach to extend the use of an already trained segmentation model from a monoplex IHC domain to a duplex IHC domain. By introducing an auxiliary CycleGAN between the duplex IHC domain and an auxiliary IF domain, we effectively resolve the ambiguous mapping between the monoplex and duplex IHC domains. This enables direct translation from the duplex IHC domain to the monoplex IHC domain. Our experiments demonstrate that proposed approach outperforms baseline methods based on color deconvolution and standard CycleGAN-based stain translation. In future, we plan to condition the generators and adapt the restaining function to different duplex IHC domains, further enhancing the flexibility and applicability of our method.

\section{Acknowledgments}
All authors are employees of AstraZeneca and do not have other relevant financial or non-financial interests to disclose.  

\section{Compliance with ethical standards}
The study was conducted in adherence with the International Council for Harmonization Good Clinical Practice guidelines, the Declaration of Helsinki, and local regulations on the conduct of clinical research.

\bibliographystyle{IEEEbib}
\bibliography{isbi-bibliography}

\end{document}